# Evaluating Temperature Scaling Calibration Effectiveness for CNNs under Varying Noise Levels in Brain Tumour Detection

*Ankur* Chanda[1], *Kushan* Choudhury[1], *Shubhrodeep* Roy[1] ,*Shubhajit* Biswas[1] and *Somenath* Kuiry[1]

[1]Department of CSE(AIML), Institute of Engineering & Management, Kolkata, India

**Abstract.** Precise confidence estimation in deep learning is vital for high-stakes fields like medical imaging, where overconfident misclassifications can have serious consequences. This work evaluates the effectiveness of Temperature Scaling (TS), a post-hoc calibration technique, in improving the reliability of convolutional neural networks (CNNs) for brain tumor classification. We develop a custom CNN and train it on a merged brain MRI dataset. To simulate real-world uncertainty, five types of image noise are introduced: Gaussian, Poisson, Salt & Pepper, Speckle, and Uniform. Model performance is evaluated using precision, recall, F1-score, accuracy, negative log-likelihood (NLL), and expected calibration error (ECE), both before and after calibration. Results demonstrate that TS significantly reduces ECE and NLL under all noise conditions without degrading classification accuracy. This underscores TS as an effective and computationally efficient approach to enhance decision confidence of medical AI systems, hence making model outputs more reliable in noisy or uncertain settings.

## 1       Introduction

In the current scenario, cancer detection has emerged as a vital sector in healthcare, with deep learning (DL) technologies being one of the crucial elements that assist in precision and effectiveness in the detection of diseases. Cancer is a mysterious and fear-inducing disease, or a collection of diseases, in fact, which have been plaguing multicellular life forms for over 200 million years. Reliable and accurate prediction is crucial in medical image analysis, where overconfident or wrong decisions can lead to serious clinical outcomes. Deep learning (DL) techniques, most notably convolutional neural networks (CNNs), have demonstrated high performance in the detection of brain tumors from MRI scans [2]. Unfortunately, although accurate, such models tend to be badly calibrated and issue probability estimates that are not true likelihoods [10]. In high-stakes environments like healthcare, this miscalibration erodes trust and can result in unwarranted or missed interventions [3].

---

[1] Corresponding author: <u>ankurchanda198@gmail.com</u>

Temperature Scaling (TS) has become a popular post-hoc calibration method because it is simple, closed-form optimal, and has low computational cost [5,10]. By scaling the confidence of predictions with one temperature parameter, TS brings the predicted probabilities closer to observed outcomes without degrading classification accuracy. Although successful in idealized environments, its performance under real imaging conditions has yet to be fully explored. MRI data tend to be contaminated with acquisition hardware noise, motion artifacts from patients, and ambient interference, and thus it is essential to test whether calibration methods continue to be effective in noisy conditions [2,4]. This paper examines the efficacy of TS in calibrating CNNs for classifying brain tumors under various noise scenarios. A light CNN is learned on a composite MRI data set [11] and tested prior to and post using TS on five forms of synthetic noise: Gaussian, Poisson, Salt & Pepper, Speckle, and Uniform. Performance is measured using accuracy, precision, recall, F1-score, Negative Log-Likelihood (NLL), and Expected Calibration Error (ECE). Results demonstrate that TS consistently reduces NLL and ECE across noise scenarios while preserving classification accuracy.

The remainder of this paper is organized as follows: Section II discusses related works on calibration methods and their evolution beyond traditional Temperature Scaling. Section III presents the proposed CNN architecture, dataset details, temperature scaling methodology, and the noise injection strategy. Section IV provides detailed experimental results and performance comparisons under different noise conditions. Finally, Section V concludes the paper with insights into findings and directions for future research.

## 2      Related works

Calibration of deep neural networks (DNNs) has gained increasing attention, particularly in safety-critical domains such as medical imaging [2,3]. Although CNNs can attain high classification accuracy, they tend to be poorly calibrated and generate overconfident predictions, which diminish trustworthiness in real-world practice [10]. Post-hoc calibration techniques are intended to address this disparity by making predicted confidence reflect true correctness.

The most common calibration method due to its simplicity, closed-form optimization, and low computational expense is Temperature Scaling (TS) [5,10]. Nonetheless, TS is weak in distribution shifts or composition tasks like semantic segmentation. In response to these setbacks, various variants have been proposed. Local Temperature Scaling (LTS) adds pixel-level calibration to TS for spatially aware tasks [6], whereas Attended Temperature Scaling (ATS) enhances robustness on small or noisy validation sets by placing greater emphasis on strong samples [8]. Adaptive Temperature Scaling (ATS/Adaptive TS) and entropy-based methods further extend calibration by including sample-level adaptations and uncertainty estimations [7,9].

While these advances have been made, most previous research measures calibration against fairly clean data sets, typically avoiding the noisy and artifact-ridden settings present in typical clinical medical imaging [2,4]. This raises the question of whether light calibration techniques like TS remain accurate when MRI scans are corrupted. The present study fills this void directly by examining TS systematically across five levels of noise to test its resilience for brain tumor identification.

# 3     Proposed method

## 3.1     CNN Architecture

In order to conduct binary classification of brain tumor images, we constructed a lightweight Convolutional Neural Network (CNN) optimized for performance and efficiency. The network structure includes four convolutional layers, which are each accompanied by ReLU activation and max-pooling operations to gradually extract spatial hierarchies as well as decrease dimensionality. Two fully connected layers are incorporated after the convolutional blocks for projecting learned features into the classification space. Dropout regularization is used between dense layers to prevent overfitting. The output layer creates raw logits, which are converted into calibrated probability scores after training via temperature scaling. The architecture is optimized to balance computational expense and classification accuracy and is ideal for resource-constrained applications, like edge devices in health care. We have used general data augmentation methods such as adding noise, random intensity and color modification, and linear corrections.

The model was trained with a batch size of 64 and a learning rate of 0.0003 for 60 epochs. For preventing overfitting, a dropout of 0.25 was used between the fully connected layers. The dataset was divided into 80% for training purposes and 20% to test generalization performance. Training was executed on the PyTorch deep learning system on a high-performance NVIDIA RTX 3050 GPU, with support from 16 GB of system memory.

## 3.2     Dataset

The model was trained using a composite dataset compiled from three publicly available repositories: Figshare, SARTAJ brain tumor dataset, and Br35H (Brain Tumor Detection 2020) from Kaggle [11]. These datasets altogether offer T1-weighted brain MRI images classified into two categories: tumor and normal. To maintain consistency, all images were resized to 256 x 256 x 3 resolution and processed with preprocessing operations such as intensity normalization and margin removal. The data was split into training, validation, and test sets randomly using stratified sampling to maintain class balance in subsets.

## 3.3     Temperature Scaling Calibration

Temperature Scaling (TS) is a post-hoc calibration technique to adjust the predicted confidence of a classifier without impacting its accuracy. Here, the logits (pre-softmax outputs) of the trained CNN are divided by a scalar temperature parameter (T), which is trained through minimizing Negative Log Likelihood (NLL) on an independent validation set. The softmax probabilities are computed as:

$$P(\hat{y}) = \frac{e^{z/f(T)}}{\sum_{j}^{n} e^{z/f(T)}} \tag{1}$$

where ($z$) represents the logit vector and ($T$) is the temperature parameter optimized via gradient-based methods A higher ($T$) leads to softer probabilities, helping align model confidence with true likelihoods. TS was applied to the CNN's output on the test set, both in clean and noisy conditions, to assess its robustness.

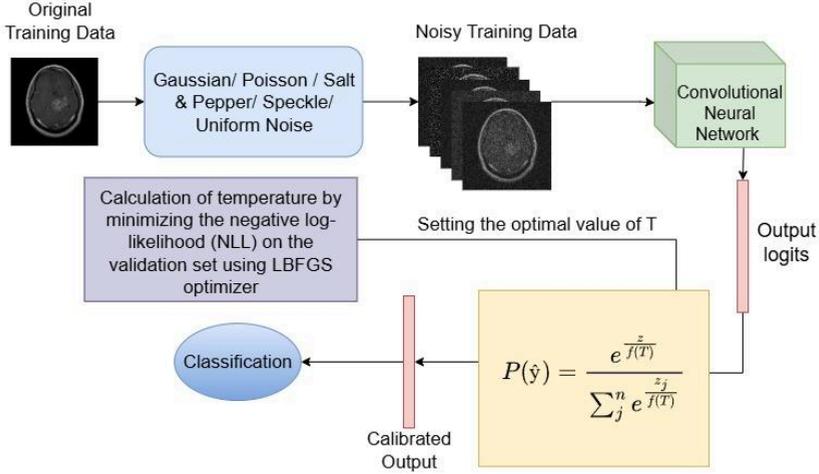

**Fig. 1.** Process Flowchart

Figure 1 depicts the workflow of the proposed approach, beginning with MRI preprocessing and CNN-based feature extraction. Synthetic noise is injected into the test data to simulate real-world conditions, followed by temperature scaling to calibrate prediction confidence. The pipeline concludes with performance evaluation across multiple metrics before and after calibration.

### 3.4 Noise Injection Strategy

To simulate real-world uncertainties in medical imaging, we introduced five types of synthetic noise to the test dataset:

#### 3.4.1 Gaussian Noise

Statistical noise known as normal noise has a probability density function matching that of the normal distribution, sometimes referred to as the Gaussian distribution. Its mean (μ) and variance ($\sigma^2$) define it. The Gaussian noise model is frequently employed in practice and even in circumstances where they are marginally relevant at best because of its mathematical simplicity.

$$P(x) = \frac{1}{\sigma\sqrt{2\pi}} e^{-\frac{(x-\mu)^2}{2\sigma^2}} \quad (2)$$

#### 3.4.2 Impulse Noise

If $b > a$, intensity $b$ will appear as a light dot in the image. Conversely, level $a$ will appear like a black dot in the image. Therefore, the presence of white and black dots in the picture mimics salt-and-pepper particles, thus also known as **salt-and-pepper noise**. Unipolar noise is when either $P_a$ or $P_b$ is zero. Quick transients like improper switching in cameras or other similar situations cause impulse noise.

$$p(z) = \begin{cases} P_a & \text{for } z = a \\ P_b & \text{for } z = b \\ 0 & \text{otherwise} \end{cases} \quad (3)$$

### 3.4.3 Poisson Noise

Also called photon noise, Poisson noise is a kind of noise resulting from the image sensor's discrete character of photons and electrons. Its prevalent in low-light situations and depends on the signal intensity.

$$P(x) = \frac{\lambda^x e^{-\lambda}}{x!} \tag{4}$$

where $P(x)$ is the likelihood of seeing $x$ photons or electrons, and $\lambda$ is the mean number of photons or electrons (mean of the distribution).

### 3.4.4 Speckle Noise

Usually affecting coherent imaging systems like radar, ultrasound, and laser, speckle noise is a multiplicative noise. The interference of coherent waves reflected from several surfaces causes this noise, which gives the picture a grainy look.

A natural characteristic of medical ultrasound imaging, speckle noise usually lowers the image quality and contrast, hence lowering the diagnostic utility of this imaging tool.

$$I_{noisy}(x, y) = I_{clean}(x, y) \cdot N(x, y) \tag{5}$$

where $I_{noisy}(x, y)$ is the noisy picture, $I_{clean}(x, y)$ is the original clean image, and $N(x, y)$ is a noise term that follows a multiplicative model, often modeled as a Rayleigh or Gamma distribution.

### 3.4.5 Uniform Noise

Uniform noise is noise that follows a uniform distribution, hence every value within a given range is equally probable.

$$P(x) = \frac{1}{b-a}, a \leq x \leq b \tag{6}$$

where $P(x)$ is the likelihood of seeing $x$ photons or electrons, and $\lambda$ is the mean number of photons or electrons (mean of the distribution).
where $P(x)$ is the probability density function.
Respectively, $a$ and $b$ are the lowest and maximum range values.

## 4   Results and discussion

### 4.1   Experimental Setup

Python was used to implement all models; Pytorch was the main deep learning framework. Training was done on a machine with 16 GB RAM and an NVIDIA RTX 3050 GPU. Using the Adam optimizer with a learning rate of $3e^{-4}$, a batch size of 64, and a maximum of 60 epochs, the CNN model was trained. Overfitting was avoided by early stopping. To improve generalization, dropout layers were applied at a 0.25 rate.

Using the L-BFGS optimization technique, the best temperature parameter ($T$) was found by reducing the Negative Log-Likelihood (NLL) loss. The unmodified test set and noise-corrupted test sets were used to assess all calibration studies.

## 4.2 Evaluation Metrics

The following criteria were used to evaluate the model's performance:
- Accuracy: Total percentage of accurate forecasts

$$Accuracy = \frac{TP+TN}{TP+TN+FP+FN} \quad (7)$$

- Precision: Ratio of actual positive forecasts among all positive forecasts

$$Precision = \frac{TP}{TP+FP} \quad (8)$$

- Recall: Percentage of true positive forecasts among all actual positives.

$$Recall = \frac{TP}{TP+FN} \quad (9)$$

- F1-Score: Harmonic mean of precision and recall.

$$F1-Score: \frac{2 \cdot (Precision \cdot Recall)}{(Precision+Recall)} \quad (10)$$

- Negative Log-Likelihood (NLL): Assesses how closely the projected probability distribution matches actual labels.

$$NLL = -\sum_{i=1}^{n} log(\hat{\pi}(y_i|x_i)) \quad (11)$$

- Expected Calibration Error (ECE): Measures the difference between projected confidence and actual accuracy across bins.

$$ECE = \sum_{m=1}^{M} \frac{|B_m|}{n} |acc(B_m) - conf(B_m)| \quad (12)$$

## 4.3 Results without Noise (Baseline)

Table 1. Baseline model performance before and after calibration

| Strategy | Precision | Recall | f1-score | Accuracy | NLL | ECE |
|---|---|---|---|---|---|---|
| Before Calibration | 0.98 | 0.975 | 0.98 | 0.98 | 0.079 | 0.016 |
| After Calibration | 0.9875 | 0.975 | 0.975 | 0.98 | 0.077 | 0.017 |

The model demonstrated strong performance prior to the introduction of noise. Remarkably, the number of false positives was 8 both prior to and after calibration. However, as seen in Fig. 2, the number of false negatives increased somewhat, from 11 to 13.

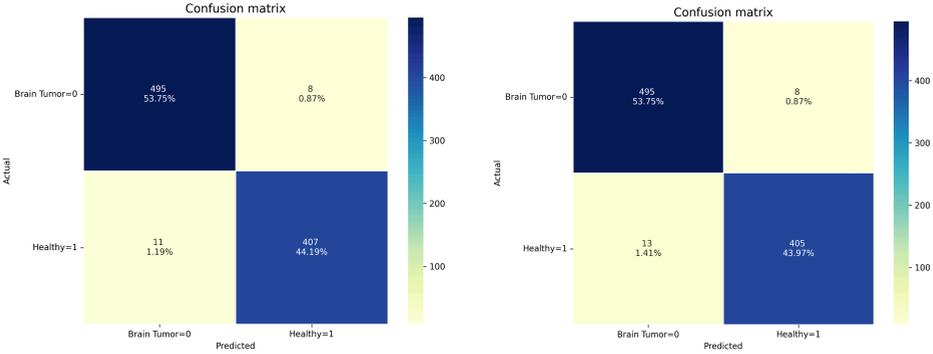

**Fig. 2.** Confusion matrix of the baseline model before & after temperature

Application of temperature scaling does not improve performance metrics like accuracy, F1-score, recall, and precision; rather, it leads to a minor reduction in these values, as shown in Table I. The small dataset size can be the cause of this effect since it might limit the efficiency of post-hoc calibration techniques, such as temperature scaling. Moreover, the complex model structure might already effectively capture the underlying data distribution in a way that additional calibration would be redundant or even detrimental in this particular case.

### 4.4 Calibration Performance under Various Noise Conditions

**Table 2.** Comparing calibration performance under Gaussian noise.

| Metric | Gaussian Noise | | | | | | |
|---|---|---|---|---|---|---|---|
| | μ=0.0, σ=0.02 | μ=0.0, σ=0.05 | μ=0.0, σ=0.1 | μ=0.0, σ=0.2 | μ=0.1, σ=0.2 | μ=0.1, σ=0.3 | μ=0.2, σ=0.3 |
| Uncalibrated | | | | | | | |
| Prec | 0.98 | 0.985 | 0.97 | 0.975 | 0.965 | 0.96 | 0.965 |
| Rec | 0.98 | 0.98 | 0.965 | 0.97 | 0.965 | 0.955 | 0.95 |
| F1 | 0.98 | 0.985 | 0.965 | 0.975 | 0.965 | 0.955 | 0.965 |
| TP | 490 | 497 | 496 | 494 | 488 | 485 | 477 |
| FN | 13 | 6 | 7 | 9 | 15 | 18 | 26 |
| FP | 3 | 9 | 23 | 15 | 17 | 20 | 16 |
| TN | 415 | 409 | 395 | 403 | 401 | 398 | 402 |
| NLL | 0.065 | 0.075 | 0.127 | 0.085 | 0.093 | 0.069 | 0.108 |
| ECE | 0.010 | 0.012 | 0.021 | 0.019 | 0.013 | 0.017 | 0.020 |
| Calibrated | | | | | | | |
| Prec | 0.985 | 0.985 | 0.955 | 0.975 | 0.975 | 0.975 | 0.98 |
| Rec | 0.98 | 0.99 | 0.955 | 0.975 | 0.97 | 0.975 | 0.955 |
| F1 | 0.98 | 0.985 | 0.955 | 0.975 | 0.975 | 0.975 | 0.965 |
| TP | 494 | 496 | 488 | 497 | 490 | 490 | 479 |
| FN | 9 | 7 | 15 | 6 | 13 | 13 | 24 |
| FP | 7 | 6 | 23 | 18 | 12 | 12 | 10 |
| TN | 411 | 412 | 395 | 400 | 406 | 406 | 408 |
| NLL | 0.060 | 0.061 | 0.113 | 0.076 | 0.089 | 0.067 | 0.093 |
| ECE | 0.007 | 0.006 | 0.009 | 0.013 | 0.011 | 0.014 | 0.013 |

| Metric | Gaussian Noise | | | | | | |
|---|---|---|---|---|---|---|---|
| | μ=0.0, σ=0.02 | μ=0.0, σ=0.05 | μ=0.0, σ=0.1 | μ=0.0, σ=0.2 | μ=0.1, σ=0.2 | μ=0.1, σ=0.3 | μ=0.2, σ=0.3 |
| | Uncalibrated | | | | | | |
| Prec | 0.98 | 0.985 | 0.97 | 0.975 | 0.965 | 0.96 | 0.965 |
| Rec | 0.98 | 0.98 | 0.965 | 0.97 | 0.965 | 0.955 | 0.95 |
| F1 | 0.98 | 0.985 | 0.965 | 0.975 | 0.965 | 0.955 | 0.965 |
| TP | 490 | 497 | 496 | 494 | 488 | 485 | 477 |
| FN | 13 | 6 | 7 | 9 | 15 | 18 | 26 |
| FP | 3 | 9 | 23 | 15 | 17 | 20 | 16 |
| TN | 415 | 409 | 395 | 403 | 401 | 398 | 402 |
| NLL | 0.065 | 0.075 | 0.127 | 0.085 | 0.093 | 0.069 | 0.108 |
| ECE | 0.010 | 0.012 | 0.021 | 0.019 | 0.013 | 0.017 | 0.020 |
| | Calibrated | | | | | | |
| Prec | 0.985 | 0.985 | 0.955 | 0.975 | 0.975 | 0.975 | 0.98 |
| Rec | 0.98 | 0.99 | 0.955 | 0.975 | 0.97 | 0.975 | 0.955 |
| | Optimal Temperature | | | | | | |
| Optimal Temp | 1.499 | 1.581 | 1.532 | 1.501 | 1.499 | 1.261 | 1.540 |

### 4.4.1 Gaussian Noise

The comparison between various Gaussian noise conditions (Table 2) highlights that calibration significantly improves the NLL and ECE scores without affecting the prediction performance metrics such as precision, recall, and F1-score. These metrics remain consistently high across all noise conditions; however, the best calibrated scores are observed when $\mu = 0.0$, $\sigma = 0.02$, showing the best precision-recall trade-off and optimal calibration performance with an optimal temperature of 1.499. The optimal temperature values range between 1.261 and 1.581 depending on the level of Gaussian noise. The lowest optimal temperature of 1.261 corresponds to $\mu = 0.1, \sigma = 0.3$, indicating that more conservative smoothing was necessary under more extreme noise conditions. Conversely, the highest optimal temperature of 1.581 corresponds to $\mu = 0.0$, $\sigma = 0.05$, suggesting that greater softening of prediction confidence was required in this configuration.

### 4.4.2 Poisson Noise and Salt & Pepper Noise

**Table 3.** Comparing calibration performance under Poisson and Salt & Pepper noise.

| Metric | Poisson | | | Salt & Pepper | | |
|---|---|---|---|---|---|---|
| | scale = 0.5 | scale = 1 | scale = 2 | salt_prob = 0.02 pepper_prob = 0.02 | salt_prob = 0.1 pepper_prob = 0.1 | salt_prob = 0.2 pepper_prob = 0.2 |
| | Uncalibrated | | | | | |
| Prec | 0.97 | 0.975 | 0.975 | 0.965 | 0.945 | 0.97 |
| Rec | 0.965 | 0.975 | 0.98 | 0.96 | 0.945 | 0.965 |
| F1 | 0.97 | 0.975 | 0.975 | 0.965 | 0.945 | 0.965 |
| TP | 485 | 495 | 491 | 485 | 478 | 493 |
| FN | 18 | 8 | 12 | 18 | 25 | 10 |
| FP | 11 | 13 | 10 | 16 | 23 | 19 |
| TN | 407 | 405 | 408 | 402 | 395 | 399 |
| NLL | 0.096 | 0.067 | 0.082 | 0.139 | 0.149 | 0.138 |
| ECE | 0.010 | 0.010 | 0.011 | 0.02 | 0.009 | 0.018 |
| Calibrated | | | | | | |
| Prec | 0.985 | 0.98 | 0.98 | 0.955 | 0.95 | 0.975 |
| Rec | 0.97 | 0.98 | 0.98 | 0.96 | 0.95 | 0.97 |
| F1 | 0.975 | 0.98 | 0.98 | 0.955 | 0.95 | 0.975 |
| TP | 486 | 496 | 493 | 482 | 477 | 490 |
| FN | 17 | 7 | 10 | 21 | 26 | 13 |
| FP | 8 | 10 | 10 | 18 | 20 | 12 |
| TN | 410 | 408 | 408 | 400 | 398 | 406 |
| NLL | 0.091 | 0.065 | 0.076 | 0.131 | 0.145 | 0.129 |
| ECE | 0.012 | 0.010 | 0.011 | 0.013 | 0.017 | 0.017 |
| Optimal Temperature | | | | | | |
| Optimal Temp | 1.435 | 1.382 | 1.452 | 1.469 | 1.399 | 1.499 |

For Poisson noise, the configuration with scale = 1 is found to be optimal in terms of balancing prediction quality and calibration, achieving an optimal temperature of 1.382 (refer to Table 3). In the case of Salt & Pepper noise, calibration performance is well-maintained even as noise levels increase. The configuration with salt_prob = 0.2 and pepper_prob = 0.2 yields the highest F1-score and the lowest NLL (0.126) among its group, at an optimal temperature of 1.499.

### 4.4.3 Uniform and Speckle noise

Table 4. Comparing calibration performance under Speckle & Uniform noise.

| Metric | Speckle scale = 0.01 | Speckle scale = 0.05 | Speckle scale = 0.1 | Uniform scale = 0.02 | Uniform scale = 0.05 | Uniform scale = 0.1 |
|---|---|---|---|---|---|---|
| Uncalibrated | | | | | | |
| Prec | 0.98 | 0.97 | 0.98 | 0.98 | 0.975 | 0.98 |
| Rec | 0.985 | 0.97 | 0.98 | 0.98 | 0.975 | 0.98 |
| F1 | 0.98 | 0.97 | 0.98 | 0.98 | 0.975 | 0.98 |
| TP | 491 | 493 | 493 | 495 | 495 | 491 |
| FN | 12 | 10 | 10 | 8 | 8 | 12 |
| FP | 5 | 18 | 9 | 8 | 13 | 7 |
| TN | 413 | 400 | 409 | 410 | 405 | 411 |
| NLL | 0.067 | 0.058 | 0.082 | 0.062 | 0.058 | 0.081 |
| ECE | 0.007 | 0.007 | 0.012 | 0.001 | 0.009 | 0.008 |
| Calibrated | | | | | | |

| Prec | 0.98 | 0.975 | 0.98 | 0.985 | 0.985 | 0.98 |
|---|---|---|---|---|---|---|
| Rec | 0.98 | 0.975 | 0.98 | 0.98 | 0.98 | 0.985 |
| F1 | 0.98 | 0.975 | 0.98 | 0.985 | 0.985 | 0.98 |
| TP | 491 | 494 | 493 | 495 | 495 | 491 |
| FN | 12 | 9 | 10 | 8 | 8 | 12 |
| FP | 7 | 13 | 9 | 7 | 7 | 6 |
| TN | 411 | 405 | 409 | 411 | 411 | 412 |
| NLL | 0.064 | 0.06 | 0.07 | 0.054 | 0.055 | 0.082 |
| ECE | 0.006 | 0.009 | 0.006 | 0.007 | 0.007 | 0.015 |
| Optimal Temperature | | | | | | |
| Optimal Temp | 1.399 | 1.175 | 1.564 | 1.501 | 1.425 | 1.314 |

The results for Uniform and Speckle noise are shown in Table 4. From the table we can see that, Temperature scaling consistently reduces NLL and ECE, even with increasing noise levels; maintains or slightly improves F1-scores, and suggests that lower optimal temperatures are sufficient under light noise levels (scale = 0.01-0.05), whereas higher temperatures (above 1.5) are necessary under heavier noise conditions.

### 4.5 Further Analysis

We extended our experiments to a more robust architecture, ResNet50, using the most optimal noise parameters identified from earlier experiments to test the effectiveness of TS on deeper models. The results are summarised in Table 5 below.
.

Table 5. Comparing calibration performance of ResNet50 under optimal noise parameters

| Metric | Gaussian ($\mu$=0.0, $\sigma$=0.02) | Poisson (scale=1) | Salt & Pepper (salt=0.2, p=0.2) | Speckle (scale=0.05) | Uniform (scale=0.05) |
|---|---|---|---|---|---|
| Uncalibrated | | | | | |
| Prec | 0.995 | 0.995 | 0.99 | 0.99 | 0.99 |
| Rec | 0.99 | 0.995 | 0.99 | 0.99 | 0.99 |
| F1 | 0.99 | 0.99 | 0.99 | 0.99 | 0.99 |
| TP | 497 | 501 | 499 | 495 | 500 |
| FN | 6 | 2 | 4 | 8 | 3 |
| FP | 3 | 2 | 5 | 4 | 5 |
| TN | 415 | 417 | 413 | 414 | 413 |
| NLL | 0.027 | 0.015 | 0.026 | 0.024 | 0.017 |
| ECE | 0.008 | 0.004 | 0.007 | 0.006 | 0.007 |
| Calibrated | | | | | |
| Prec | 0.995 | 0.995 | 0.99 | 0.99 | 1.00 |
| Rec | 0.995 | 0.995 | 0.99 | 0.99 | 1.00 |
| F1 | 0.99 | 0.995 | 0.99 | 0.99 | 1.00 |
| TP | 500 | 503 | 500 | 495 | 503 |
| FN | 3 | 0 | 3 | 8 | 0 |
| FP | 1 | 2 | 3 | 5 | 0 |

| Metric | Gaussian (μ=0.0, σ=0.02) | Poisson (scale=1) | Salt & Pepper (salt=0.2, p=0.2) | Speckle (scale=0.05) | Uniform (scale=0.05) |
|---|---|---|---|---|---|
| Uncalibrated | | | | | |
| Prec | 0.995 | 0.995 | 0.99 | 0.99 | 0.99 |
| Rec | 0.99 | 0.995 | 0.99 | 0.99 | 0.99 |
| F1 | 0.99 | 0.99 | 0.99 | 0.99 | 0.99 |
| TP | 497 | 501 | 499 | 495 | 500 |
| FN | 6 | 2 | 4 | 8 | 3 |
| FP | 3 | 2 | 5 | 4 | 5 |
| TN | 415 | 417 | 413 | 414 | 413 |
| NLL | 0.027 | 0.015 | 0.026 | 0.024 | 0.017 |
| ECE | 0.008 | 0.004 | 0.007 | 0.006 | 0.007 |
| Calibrated | | | | | |
| Prec | 0.995 | 0.995 | 0.99 | 0.99 | 1.00 |
| Rec | 0.995 | 0.995 | 0.99 | 0.99 | 1.00 |
| TN | 417 | 417 | 415 | 415 | 418 |
| NLL | 0.025 | 0.014 | 0.024 | 0.025 | 0.015 |
| ECE | 0.003 | 0.003 | 0.003 | 0.006 | 0.006 |
| Optimal Temperature | | | | | |
| Optimal Temp | 1.321 | 1.290 | 1.321 | 1.316 | 1.218 |

In comparing performance, we observe that temperature scaling really improves both the false positive (FP) and false negative (FN) rates for ResNet, while leaving overall model accuracy unaffected and also reducing the NLL loss and ECE. These findings suggest that the benefits of calibration methods such as temperature scaling are not confined to simple CNN baselines, but can also enhance the performance of deeper, state-of-the-art architectures.

## 5 Conclusion

Temperature Scaling (TS) is a straightforward and efficient method for post-training calibration that improves the reliability of the model against different types of noise, i.e., Gaussian, Poisson, Salt & Pepper, Speckle, and Uniform noise. Experimental results reveal that TS greatly minimizes the miscalibration of models by achieving significant improvement in some major calibration metrics such as Negative Log-Likelihood (NLL) and Expected Calibration Error (ECE). Notably, this calibration is done without sacrificing the quality of predictions—important performance metrics such as precision, recall, F1-score, and overall accuracy are robust in all noise settings. These findings indicate that TS is a gentle but strong model confidence boost facility, especially important in sensitive tasks like medical diagnosis, where the decision must be accurate and trustworthy.

As part of our future studies,we want to investigate the performance of calibration methods such as TS and others on a variety of contemporary deep learning models on much larger datasets for generalization. Specifically, we intend to analyze how well they perform at reducing false negatives and false positives, and specifically why they are particularly valuable in high-risk domains like medicine, where both have the potential to be severely damaging. We also intend to broaden the research by comparing Temperature Scaling to other cutting-edge calibration approaches including Adaptive Temperature Scaling, Entropy-Based Temperature Scaling, and Sample-Dependent methods. Implementation of such methods will enable more thorough assessment of calibration performance, especially across changing noise levels and distributional drift in medical image data. This direction will enable further development of knowledge on how calibrated confidence scores can not only enhance better trust in predictions but also safer and more accurate outcomes.

## References

1. Siegel, Rebecca L., Angela N. Giaquinto, and Ahmedin Jemal. "Cancer statistics, (2024)." *CA: a cancer journal for clinicians* **74**, no. 1 (2024).
2. Babu Vimala, B., Srinivasan, S., Mathivanan, S. K., Mahalakshmi, Jayagopal, P., & Dalu, G. T. (2023). Detection and classification of brain tumor using hybrid deep learning models. *Scientific reports*, **13(1)**, 23029. https://doi.org/10.1038/s41598-023-50505-6
3. Pashayan, N., & Pharoah, P. D. P. (2020). The challenge of early detection in cancer. *Science* (New York, N.Y.), **368(6491)**, 589–590. https://doi.org/10.1126/science.aaz2078.
4. Ma Y, Peng Y, Wu T-Y, Balas VE. Transfer learning model for false positive reduction in lymph node detection via sparse coding and deep learning. *Journal of Intelligent & Fuzzy Systems*. (2022);43(2):2121-2133. doi:10.3233/JIFS-219312
5. Balanya, S.A., Maroñas, J. & Ramos, D. Adaptive temperature scaling for Robust calibration of deep neural networks. *Neural Comput & Applic* **36**, 8073–8095 (2024). https://doi.org/10.1007/s00521-024-09505-4
6. Ding, Zhipeng & Han, Xu & Liu, Peirong & Niethammer, Marc. (2020). Local Temperature Scaling for Probability Calibration. 10.48550/arXiv.2008.05105.
7. Joy, Tom & Pinto, Francesco & Lim, Ser-Nam & Torr, Philip & Dokania, Puneet. (2023). Sample-Dependent Adaptive Temperature Scaling for Improved Calibration. *Proceedings of the AAAI Conference on Artificial Intelligence*. 37. 14919-14926. 10.1609/aaai.v37i12.26742.
8. Mozafari, Azadeh & Gomes, Hugo & Leão Neto, Wilson & Janny, Steeven & Gagné, Christian. (2019). Attended Temperature Scaling: A Practical Approach for Calibrating Deep Neural Networks.
9. Balanya, Sergio & Maroñas, Juan & Ramos, Daniel. (2024). Adaptive temperature scaling for Robust calibration of deep neural networks. *Neural Computing and Applications*. 36. 1-23. 10.1007/s00521-024-09505-4.
10. Guo, Chuan & Pleiss, Geoff & Sun, Yu & Weinberger, Kilian. (2017). On Calibration of Modern Neural Networks. 10.48550/arXiv.1706.04599.
11. Sartaj Bhuvaji, Ankita Kadam, Prajakta Bhumkar, Sameer Dedge, and Swati Kanchan. (2020). Brain Tumor Classification (MRI) [Data set]. *Kaggle*. https://doi.org/10.34740/KAGGLE/DSV/1183165